\documentclass[11pt]{article}

\usepackage[final]{acl}

\usepackage{times}
\usepackage{latexsym}

\usepackage[T1]{fontenc}

\usepackage[utf8]{inputenc}

\usepackage{silence}
\usepackage{microtype}

\usepackage{inconsolata}

\usepackage{graphicx}

\usepackage{arydshln}

\usepackage{amsmath}
\usepackage{amssymb}
\usepackage{pifont}

\usepackage{subcaption}

\usepackage{tcolorbox}

\usepackage{booktabs}

\usepackage{multirow}

\usetikzlibrary{shapes.geometric, arrows, positioning}

\tikzset{
  model/.style = {rectangle, draw, minimum width=2.2cm, minimum height=0.9cm, font=\sffamily\tiny},
  question/.style = {rectangle, align=left, font=\sffamily\tiny, text width=6.5cm},
  arrow/.style = {->, thick, shorten <=1pt, shorten >=1pt},
  iterlabel/.style = {font=\sffamily\tiny\bfseries, anchor=north east, inner sep=1pt}
}

\usepackage{pifont}
\newcommand\blfootnote[1]{%
  \begingroup
  \renewcommand\thefootnote{}\footnote{#1}%
  \addtocounter{footnote}{-1}%
  \endgroup
}

\title{Leveraging Turn-taking Dynamics for Intent Recognition \\ in Multi-party Conversations}

\author{Galo Castillo-L\'{o}pez \quad Alexis Lombard \quad Ga\"{e}l de Chalendar \quad Nasredine Semmar \\
        Universit\'{e} Paris-Saclay, CEA, List, Palaiseau, France \\ \{\texttt{galo-daniel.castillolopez, alexis.lombard, gael.de-chalendar, nasredine.semmar}\}\texttt{@cea.fr}  }

\begin{document}
\maketitle
\begin{abstract}
We propose a multi-task learning approach for multi-party dialogue intent recognition that leverages an auxiliary task that models turn-taking dynamics. Specifically, we introduce turn-transition entropy, a self-supervised target computed from the sequence of speaker transitions, which quantifies the predictability of interaction patterns. Experiments on multiple pre-trained models demonstrate that incorporating this auxiliary task improves intent recognition performance, outperforming existing approaches which ignore multi-party interaction dynamics. We find that our proposed continuous target can be learned as a single-task objective, suggesting that it is an actual signal carrying useful information.

\end{abstract}

\section{Introduction}
\label{sec:introduction}
\blfootnote{This paper has been accepted for publication at EMNLP Industry Track 2026 and corresponds to the author's version of the work.}
Detection of users' intents is a fundamental component of task-oriented dialogue systems, enabling conversational agents to understand user needs and generate appropriate responses \cite{zhu-etal-2025-survey}. While intent recognition has been extensively studied in dyadic dialogues \cite{zawbaa-etal-2024-improved,arora-etal-2024-intent,sali-toraman-2025-navigating}, the multi-party scenario, where agents interact with multiple users simultaneously, has been overlooked. Recent work has investigated intent recognition in multi-party conversations (MPCs) using both small and large language models \cite{castillolópez2025intentrecognitionoutofscopedetection,liu2026should}. However, most approaches fail to leverage information from the inherent multi-party interaction dynamics, such as turn-taking behavior, conversation disentanglement, and others \cite{ganesh2023survey,sapkota2025multi}.

\begin{figure}[t]
\centering
\includegraphics[width=.95\linewidth]{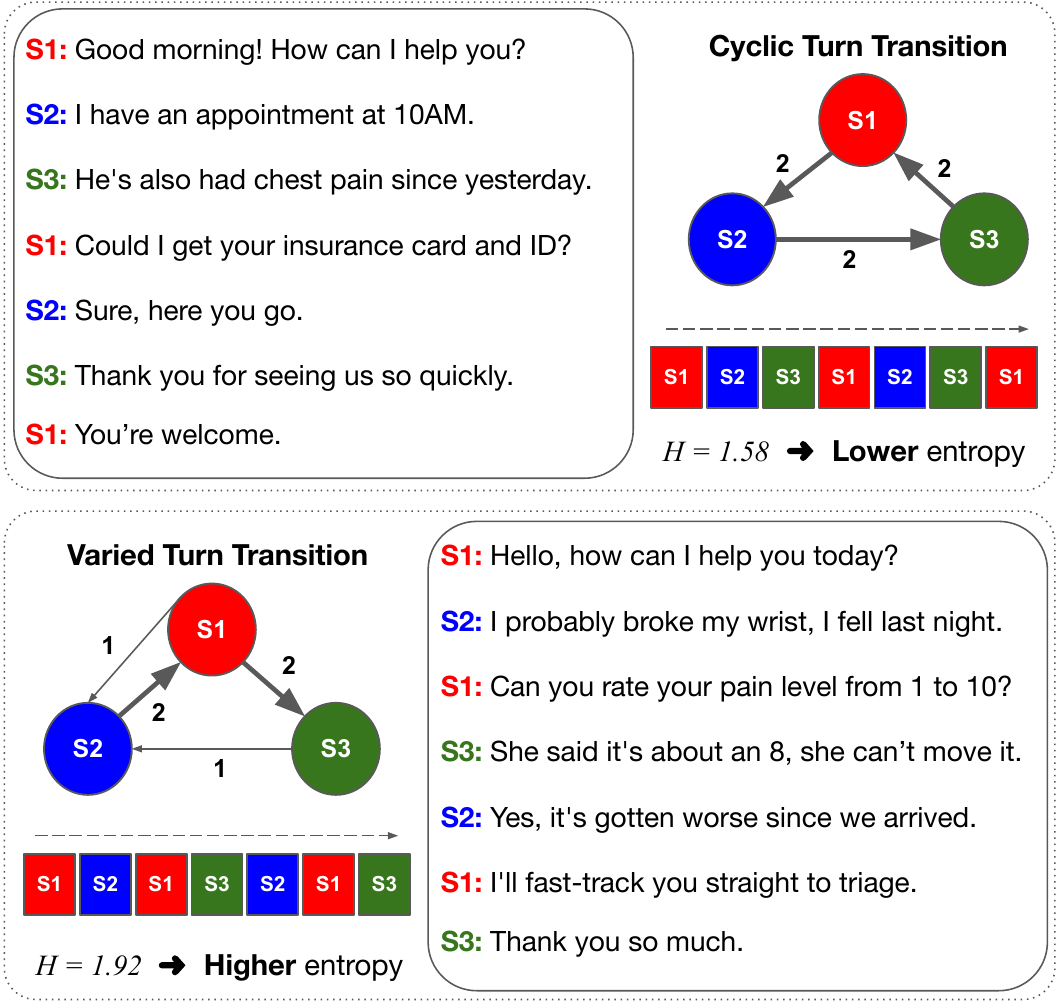}
\caption[]{Examples of low- and high-entropy turn transitions. The top conversation follows a structured pattern of speaker transitions, presenting a lower turn-transition entropy. The bottom conversation exhibits more varied and less organized speaker transitions, leading to a higher entropy\footnotemark. Note that both examples have the same number of speakers and turn shifts.}
\label{fig:examples_dialogues}
\end{figure}

In this paper, we propose a multi-task learning approach that incorporates an auxiliary task to capture turn-taking dynamics for multi-party dialogue intent recognition. Unlike dyadic conversations, where turn-taking is relatively simple to anticipate, MPCs involve multiple potential next speakers, making speaker interaction patterns more complex \cite{skantze2021turn,castillo-lopez-etal-2025-survey}. We argue that such patterns provide useful contextual cues to model MPCs and therefore improve intent recognition in these scenarios. Although tasks such as next-speaker prediction and addressee detection have been widely studied \cite{ijcai2022p768,10795688,mori-etal-2026-analysing,inoue-etal-2025-llm}, they often require additional annotations or assumptions about the number of participants.

\footnotetext{We define the entropy computation in Section \ref{sec:entropy}.}

To address these limitations, we introduce an auxiliary regression task that predicts the uncertainty of turn-taking patterns in the dialogue context. Given the turns preceding a target utterance, we extract speaker-transition pairs (e.g., Speaker A $\rightarrow$ Speaker B, Speaker B $\rightarrow$ Speaker C) and compute their Shannon entropy \cite{shannon1948mathematical}. This \textbf{turn-transition entropy} quantifies the predictability of interaction dynamics: low entropy indicates more organized turn-taking, whereas high entropy shows more unpredictable turn transitions (Figure~\ref{fig:examples_dialogues}). By learning to predict this quantity, intent recognition models are encouraged to capture interaction dynamics inherent to MPCs without requiring additional annotations or a fixed number of participants. Our main contributions are as follows:

\begin{itemize}
\item We introduce \textbf{turn-transition entropy}, a self-supervised continuous target that captures turn-taking dynamics in multi-party dialogues.
\item We evaluate the practical use of the turn-transition entropy as an auxiliary task for fine-tuning multi-party intent recognition models.
\end{itemize}

\section{Related Work}
\label{sec:related_work}
Intent detection has been widely studied in dyadic dialogues \cite{wang-etal-2023-app,gautam-etal-2024-class,benayas2025comparative}. More recently, the use of LLMs for intent recognition has gained attention in single-turn \cite{sali-toraman-2025-navigating,castillo-lopez-etal-2026-ddair,arora-etal-2024-intent} and multi-turn dialogue \cite{liu2026balancing}. Multitask learning methods have proposed incorporating auxiliary tasks such as named entity recognition \cite{perdana2025multi,benayas2021unified}, dialogue act classification  \cite{firdaus2021deep}, and slot filling \cite{he2021multitask,firdaus2023multitask}. These multitask methods, however, require additional annotations that are not feasible in many real-world applications. \citet{castillolópez2025intentrecognitionoutofscopedetection} presented a hybrid approach that combines small and large language models in few-shot settings on MPCs. Multimodal fusion methods in MPCs have also been investigated \cite{chen2024dual,shen2025mess}. However, none of such works explicitly aim to capture inherent multi-party behavior.


Research in dialogue act classification, a similar dialogue multiclass classification task at utterance level, has explored modeling multi-parity \cite{anwar2024dual}. Most studies have investigated modeling speaker-related dynamics for improving dialogue understanding \cite{fu2025ham}. \citet{he-etal-2021-speaker-turn} addressed the task by introducing speaker-turn embeddings added to utterance representations before contextual encoding. This mechanism efficiently signals speaker switch or continuation, but reducing speaker identities to a binary pattern comes at the expense of retaining speaker-specific turn taking information. This limitation, most consequential in MPCs, was addressed by \citet{qamar-etal-2023-speaking}, who proposed a speaker-specific graph model for multi-party dialogue act classification. Their approach learns graph-structured speaker representations from speaker–utterance connections and concatenates them with contextual utterance representations to model local speaker behavior. However, this makes speaker modeling dependent on an explicit speaker–utterance graph neural module for each dialogue. Additionally, it primarily captures local speaker behavior rather than the global distributional structure of turn transitions. 

In contrast to prior research, we incorporate turn-taking dynamics for multi-party intent recognition through a self-supervised multitask strategy. That is, without requiring additional costly annotations. In addition, our approach does not rely on a fixed number of speakers. Rather than using speaker-turn embeddings or graph-derived speaker representations, we introduce turn-transition entropy as an auxiliary regression signal, encouraging transformer-based models to capture the structure of speaker transitions while optimizing the main intent classification objective.

\section{Experimental Procedure}
\label{sec:experiments}
Let $\mathcal{D}$ be a dataset of multi-party dialogues, where each dialogue $d_k = \{(u_i, p_i, y_i)\}_{i=1}^{n_k}$ is a sequence of $n_k$ utterances, with $u_i$ denoting the $i$-th utterance, $p_i \in \mathcal{P}_k$ its speaker from a dialogue-specific set of participants $\mathcal{P}_k$, and. $y_i \in \mathcal{Y}$ its intent label. $\mathcal{Y} = \{1, \dots, m\}$ denotes the predefined set of $m$ intent labels. Our aim is to build a multiclass classification system that assigns each utterance $u_i$ to its corresponding intent $y_i \in \mathcal{Y}$. Our contribution is to leverage the conversational interactions between speakers in  $\mathcal{P}_k$.

\subsection{Utterance Modeling}
\label{sec:utterance_modeling}
We use three different encoders to represent the text from utterances: BERT \cite{devlin2019bert}, RoBERTa \cite{liu2019roberta}, and DeBERTaV3 \cite{he2021debertav3}. We use the large versions of all pre-trained models. More details of the models we use are included in Appendix~\ref{app:language_models}. In addition to the utterance to classify and in line with previous work, we prepend the encoded representation of previous utterances to include context information. 

\subsection{Context Information Modeling}
\label{sec:speaker_turn_modeling}
\begin{figure*}[t]
\centering
\includegraphics[width=.65\linewidth]{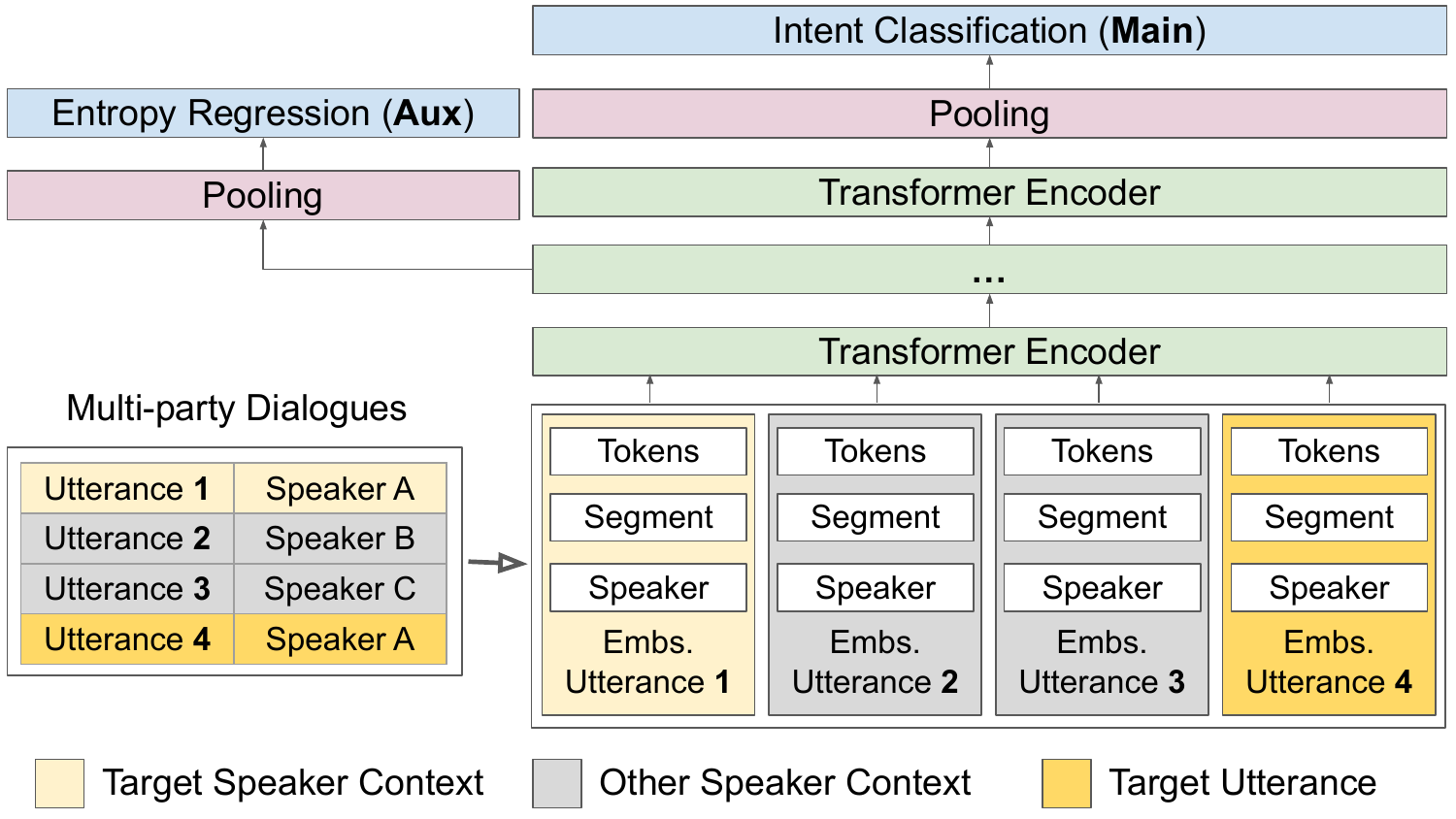}
\caption[]{Overview of our multitask learning approach architecture. Input representation includes model token embeddings; segment IDs to distinguish contextual from target utterances (i.e. latest turn utterance, to classify into any intent); and speaker embeddings to discriminate context uttered by the speaker producing the target utterance.}
\label{fig:overview_architecture}
\end{figure*}

We add additional learnable embeddings to our encoded text representation to incorporate context information. Following \cite{zhang2024mintrec2}, we add segment embeddings to let the model discriminate between contextual utterances and the target utterance $u_t$ to classify. In \cite{zhang2024mintrec2}, only utterances previously produced by the target speaker $p_t$ are prepended as context. However, we find that performance improves when incorporating all previous utterances as context, regardless of their speaker. To compensate for the loss of speaker identity that comes with this broader context, we introduce an additional speaker embedding that distinguishes utterances produced by the target speaker $p_t$ from those produced by other participants. Experiments showing the advantages of this modification are described in Appendix~\ref{app:speaker_emb_experiments}. The final input representation and an overview of our architecture is illustrated in Figure~\ref{fig:overview_architecture}.

\subsection{Turn-Transition Entropy Auxiliary Task}
\label{sec:entropy}
In multi-party dialogues, the way participants alternate in taking turns may carry implicit structural information about the nature of the conversation. To capture this, we introduce the \textbf{turn-transition entropy}, a continuous measure that characterizes the distribution of speaker transitions in a dialogue. Given a dialogue $d_k = \{(u_i, p_i, y_i)\}_{i=1}^{n_k}$, we define the set of observed turn-transition pairs for a target utterance $u_t$:
\begin{equation}
    \mathcal{A}_t = \{(p_{i-1}, p_i) \mid i = 1, \dots, t\}
\end{equation}

where each pair $(p_{i-1}, p_{i})$ represents a transition from the speaker of utterance $u_{i-1}$ to the speaker of utterance $u_{i}$. Let $\mathcal{A}_t^*$ denote the set of unique transition pairs observed in $\mathcal{A}_t$, and $c(a)$ denote the number of times pair $a \in \mathcal{A}_t^*$ occurs in $\mathcal{A}_t$. The probability of each observed turn shift is estimated as:
\begin{equation}
    P(a) = \frac{c(a)}{|\mathcal{A}_t|}
\end{equation}
The turn-transition entropy for target utterance $u_t$ is then defined as the Shannon entropy \cite{shannon1948mathematical} over the observed transition pairs:
\begin{equation}
    H_t = -\sum_{a \in \mathcal{A}_t^*} P(a) \log P(a)
\end{equation}

\begin{figure}[ht]
    \centering
    \includegraphics[width=.9\linewidth]{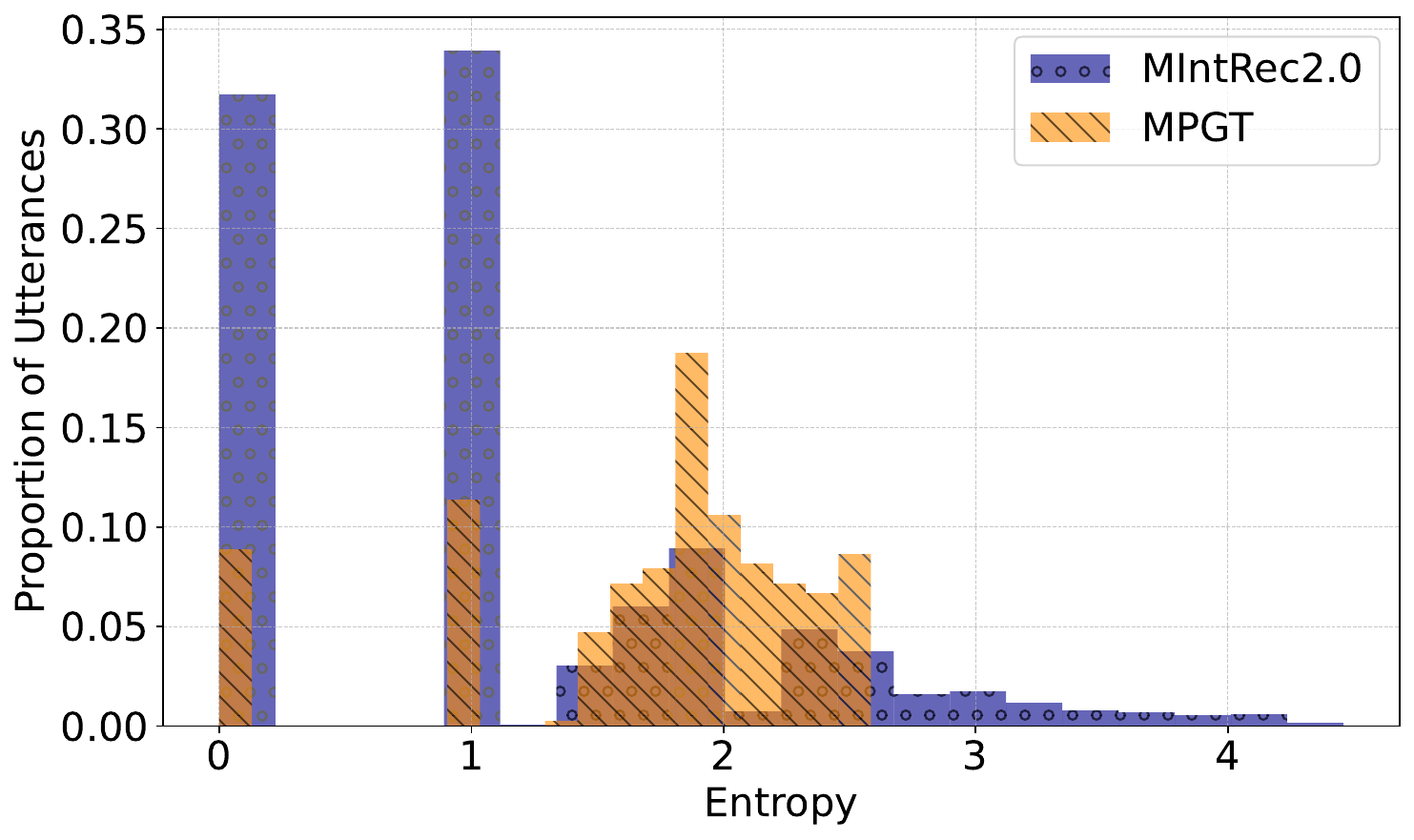}
    \caption{Entropy distributions computed on the training sets of MIntRec2.0 and MPGT.}
    \label{fig:entropy_distributions}

\end{figure}

As a result, we obtain a computed continuous value for each utterance. Unlike speaker-count or turn-count statistics, $H_t$ reflects how evenly distributed the turn-taking transitions are among the pairs observed until turn $t$. A low entropy indicates an organized order of turn transitions (i.e. a few transitions dominate), while a high entropy reflects a more diverse set of turn transitions, i.e. more unpredictable speaker-wise turn shifts. We propose incorporating the defined entropy value as an auxiliary regression target in a multitask learning framework alongside the main intent classification task, to leverage the multi-party nature of the conversation as an additional source of information, encouraging the model to capture turn-taking dynamics to improve intent classification. The distributions of the computed entropy values are shown in Figure \ref{fig:entropy_distributions}.

Finally, we argue that the proposed turn-transition entropy variable may correlate with utterance properties such as turn index, context length, and speaker count. This could make unclear whether the model captures turn taking predictability or simpler dialogue properties. Therefore, we propose a normalized version of the auxiliary variable. In particular, the \textbf{normalized turn-transition entropy} variable is obtained by dividing the actual entropy values by the speaker counts observed in the context utterances. Consequently, we conduct our multitask learning experiments on both: the turn-transition entropy and the normalized turn-transition entropy variables. In Appendix~\ref{app:app_correlation_analysis}, we perform a relationship analysis between the auxiliary targets and the utterance properties that we enumerated previously.

\subsection{Intent Classification}
\label{sec:classification}
We consider two learning approaches for the intent classification task: a single-task (our baseline) and a multitask setting.

\paragraph{Single-Task.} In the single-task setting, a classification head consisting of a linear layer followed by a softmax activation is placed on top of the language model to predict the intent label $y_t \in \mathcal{Y}$. The model is optimized using the Cross-Entropy Loss $\mathcal{L}_{\text{cls}}$. Models are fine-tuned across 60 epochs with an early stopping mechanism. More details about model fine-tuning are provided in Appendix~\ref{app:baseline_imp}.

\paragraph{Multitask.} In the multitask setting, we additionally incorporate the turn-transition entropy $H_t$ as an auxiliary regression target. A regression head is connected to an intermediate layer of the language model, specifically the $l$-th transformer layer, where $l$ is a tunable hyperparameter. The hidden states at layer $l$ are pooled and fed into a regression head trained to predict $H_t$, optimized using the Mean Squared Error Loss $\mathcal{L}_{\text{reg}}$.
The two objectives are combined using a weighted geometric mean, which is invariant to the scale of the individual losses \cite{chennupati2019multinet++}, controlled by a hyperparameter $\lambda \in [0, 1]$ as $\mathcal{L} = \mathcal{L}_{\text{cls}}^{(1-\lambda)} \cdot \mathcal{L}_{\text{reg}}^{\lambda}$.
The hyperparameter $\lambda$ controls the relative importance of the auxiliary regression task with respect to the main classification task. These additional hyperparameters are optimized using the Tree-structured Parzen Estimator approach \cite{bergstra2011algorithms,watanabe2023tree} on Optuna\footnotemark \cite{akiba2019optuna}. Models are fine-tuned across 60 epochs with early stopping. More details about model fine-tuning are provided in Appendix~\ref{app:multitask_imp}.

\footnotetext{{https://optuna.org/}}

\subsection{Large Language Model Baseline}
\label{sec:emnlp_speaker_turn_modeling}
In addition to the encoder models fine-tuning, we conduct a baseline experiment using \texttt{DeepSeek-R1-Distill-Llama-8B} on zero-shot settings for the intent classification task, following the zero-shot classification described in \cite{castillolópez2025intentrecognitionoutofscopedetection}. Figure~\ref{fig:emnlp_llm_baseline_prompt} in Appendix~\ref{app:large_language_models} depicts the prompt template we use for the task.

\subsection{Prior Work Baselines}
\label{sec:prior_baseline}
We use the results reported in \cite{zhang2024mintrec2} as a baseline to compare with our work. Moreover, we implement the Label Space Reduction method proposed by \citet{castillolópez2025intentrecognitionoutofscopedetection}. Their two-stage approach consists in fine-tuning BERT and routing uncertain (i.e. hard) inferences to a LLM. We use \texttt{DeepSeek-R1-Distill-Llama-70B}, which provided the best results in their work. Both baseline methods are implemented on BERT\textsubscript{LARGE}.

Additionally, we implement the speaker-aware graph-based method of \cite{qamar-etal-2023-speaking} from the paper’s methodological description. Unlike the original study, our evaluation covers both RoBERTa\textsubscript{BASE} and RoBERTa\textsubscript{LARGE}. Following the original setup, the RoBERTa embedding layer and all transformer layers except the final one are frozen; only the final transformer layer is fine-tuned before training the graph component. Each dialogue chunk is represented as a speaker-utterance graph with one node per utterance and speaker; bidirectional utterance-speaker links yield ($2|D|$) directed edges with a single relation type. Two RGAT layers, each with two attention heads and hidden size 400, compute utterance-level speaker representations that are concatenated with contextual utterance representations before classification. The complete hyperparameter configuration used for reproduction is reported in Appendix~\ref{app:prior_imp}.

We further implement the speaker-turn embedding baseline of \cite{he-etal-2021-speaker-turn} from the authors' public implementation, only replacing dialogue-act targets with intent labels and RoBERTa\textsubscript{BASE} with RoBERTa\textsubscript{LARGE}. 

\subsection{Datasets}
\label{sec:datasets}
We use two public multi-party conversations corpora in this work. The first corpus is \textbf{MPGT}, which is a set of 29 dialogues between hospital visitors and a receptionist robot \cite{addlesee-etal-2023-multi}. The second corpus is \textbf{MIntRec2.0}, a multimodal dataset of 15K dialogue utterances from TV shows \cite{zhang2024mintrec2}. In this work we are interested in systems working with text input data, hence we only use the dialogue transcripts data. MPGT and MIntRec2.0 define 8 and 30 intents, respectively. Additionally, we use \textbf{MEETInt}, an internal corpus of 75 spoken MPCs from in-person professional meetings, containing 100K utterances and 12 intents. More information about the datasets is detailed in Appendix~\ref{app:dataset_info}.

\subsection{Evaluation}
\label{sec:evaluation}
We evaluate our fine-tuned intent classification models using the average macro F1 and accuracy over 5 runs with different seeds. The evaluation of the regression task is performed using the Mean Absolute Error (MAE).

\section{Results}
\label{sec:results}

\begin{table*}[t]
\small
\centering
\begin{tabular}{c l c c | c c }
 \toprule

 & & \multicolumn{2}{c|}{MPGT\textsubscript{m}} & \multicolumn{2}{c}{MIntRec2.0}\\

Model & Method & F1 & Acc. & F1 & Acc. \\ 

\midrule

\multirow{5}{*}{BERT} & CTX TEXT \footnotesize{\cite{zhang2024mintrec2}}\footnotemark & - - & - - & 53.61 & 59.04 \\

 & Label Space Reduction \footnotesize{\cite{castillolópez2025intentrecognitionoutofscopedetection}} & 71.38 & 91.73 & 41.92 & 41.66  \\

 & baseline (ours) & 88.81 & 95.19 & 55.07 & 60.33  \\

 & TT Entropy Aux. Task (ours)  & \textbf{92.80} & 96.09 & 55.87 & 60.22  \\

 & Norm. TT Entropy Aux. Task (ours)  & 92.05 & \textbf{96.69} & \textbf{56.01} & \textbf{60.35}  \\

\midrule
\midrule

\multirow{5}{*}{RoBERTa} & Turn Modeling \footnotesize{\cite{he-etal-2021-speaker-turn}} & 73.64  & 92.92 & 51.40 & 56.40 \\

 & Turn Aware Speaker Graph \footnotesize{\cite{qamar-etal-2023-speaking}} & 81.28  & 94.29 & 52.82 & 57.92 \\

 & baseline (ours) & 86.32 & 96.39 & 55.96 & 61.29  \\

 & TT Entropy Aux. Task (ours)  & 88.43 & 95.94 & 56.92 & 60.59  \\

 & Norm. TT Entropy Aux. Task (ours)  & \textbf{89.48} & \textbf{96.39} & \textbf{57.06} & \textbf{61.49}  \\

\midrule
\midrule

\multirow{3}{*}{DeBERTa} & baseline (ours) & 80.73  & 94.29 & 55.88 & 60.79 \\

 & TT Entropy Aux. Task (ours)  & 84.56 & 95.49 & 56.53 & 60.37  \\

 & Norm. TT Entropy Aux. Task (ours)  & \textbf{85.12} & \textbf{96.99} & \textbf{57.20} & \textbf{61.71}  \\

\midrule
\midrule

DeepSeek-R1 8B & zero-shot (ours) & 39.51 & 32.33 & 31.54 & 32.12 \\

\bottomrule

\end{tabular}
\caption{Performance evaluation on the intent classification task. Scores correspond to the average macro F1 and accuracy over 5 runs. Scores in \textbf{bold} highlight \textbf{the best} performance on the same backbone model.}

\label{table:main_results}
\end{table*}

\footnotetext{Performance metrics reported in the original paper.}

Table \ref{table:main_results} shows the main results of all baselines and our multitask learning approach on all pre-trained models. We observe that our baseline method presents higher macro-F1 and accuracy scores than prior work baselines on both datasets. We also observe that regardless of the backbone model, incorporating our proposed turn-transition entropy auxiliary task --normalized or unnormalized-- leads to a macro-F1 score improvement with respect to our baseline on both MPGT and MIntRec2.0. BERT, in multitask settings, achieves the largest macro F1 on MPGT, while DeBERTa shows the highest F1 on MIntRec2.0. We find that in all cases, except for BERT on MPGT, the normalized version of the turn-transition entropy variable provides the largest improvements over the baseline models. The second best F1 scores are obtained on the unnormalized approach. Furthermore, we see that performance improvements are not large on MIntRec2.0, compared to MPGT. Nonetheless, we note that macro-F1 improvements are consistent across models. Also, LLM performance is much lower than the ones obtained by our baseline and multitask methods. To the best of our knowledge, recent studies have consistently shown that fine-tuned BERT-like models generally perform better than LLMs on in-context learning settings for intent detection, as suggested by \citet{zhang2024mintrec2}

The Label Space Reduction (LSR) method \cite{castillolópez2025intentrecognitionoutofscopedetection} shows the poorest performance among all models and corpora. However, it is important to note that such a performance comparison is not fair, as the LSR method is conducted on few-shot settings. Moreover, we see that there are 3 to 5 point performance difference between our baseline and \cite{he-etal-2021-speaker-turn,qamar-etal-2023-speaking} on the two datasets. Table \ref{table:comparison_roberta} in Appendix \ref{app:comparison_roberta} shows that both methods implemented on RoBERTa\textsubscript{BASE} present a higher performance than the same implementation on RoBERTa\textsubscript{LARGE}. Nevertheless, our multitask learning approach keeps outperforming both methods. Finally, we observe that our baseline approach on BERT outperforms \cite{zhang2024mintrec2} on MIntRec2.0. We argue that such an enhancement is due to the incorporation of the speaker embedding we propose and describe in Section \ref{sec:speaker_turn_modeling}. An analysis on the improvement over \cite{zhang2024mintrec2} is described in Appendix \ref{app:speaker_emb_experiments}. Finally, in line with the results on public corpora, Table \ref{table:results_meetintent} shows that our proposed approach outperforms all baseline methods on MEETInt. We see that our improvements over works by \citet{he-etal-2021-speaker-turn} and \citet{qamar-etal-2023-speaking} are about 4 and 2 points on macro F1, respectively. 

\begin{table}[!ht]
\small
\centering
\begin{tabular}{ l  c  c }

 \toprule

Method & F1 & Acc. \\ 

\midrule
 \cite{he-etal-2021-speaker-turn} & 70.90 & 79.15 \\
 \cite{qamar-etal-2023-speaking} & 72.49 & 80.72 \\
baseline (ours) & 73.54 & 81.00  \\
TT Entropy Aux. Task (ours) & 74.31 & 81.83 \\

Norm. TT Entropy Aux. Task (ours) & \textbf{74.78} & \textbf{81.95} \\

\bottomrule

\end{tabular}
\caption{Evaluation on MEETInt with RoBERTa\textsubscript{LARGE}. Scores correspond to the average macro F1 and accuracy over 5 runs. Scores in \textbf{bold} highlight \textbf{the best} performance.}

\label{table:results_meetintent}
\end{table}


\section{Turn-Transition Entropy Learning as a Single Task}
\label{sec:entropy_singletask}
\begin{table}[ht]
\small
\centering
\begin{tabular}{ l c c c c }

 \toprule

 & \multicolumn{2}{c}{MPGT} & \multicolumn{2}{c}{MIntRec2.0}\\

Model & random & real & random & real \\ 

\midrule

BERT & 0.680 & 0.274 & 1.165 & 0.379 \\

RoBERTa & 0.677 & 0.205 & 1.161 & 0.428 \\

DeBERTa & 0.682 & 0.220 & 1.162 & 0.330 \\
\bottomrule

\end{tabular}
\caption{Mean Absolute Error (MAE) on the regression target learned as a single task on real and random entropy values. Scores correspond to the average MAE over 5 runs. A lower MAE suggests an ease on the continuous variable learning.}

\label{table:results_single_task}
\end{table}

To better understand whether turn-transition entropy constitutes a learnable signal, we isolate it from the intent classification task and evaluate it as a standalone regression problem, i.e. as a single task. We replace the classification head of our backbone models (described in Section~\ref{sec:classification}) with a regression head and fine-tune the models to predict turn-transition entropy from the text features. In addition, we compare this setting against a control condition in which entropy values are randomly shuffled across training instances. By doing so, our aim is to assess whether the model learns meaningful patterns associated with turn-transition entropy or merely fits arbitrary continuous targets.

\begin{figure}[h]
    \centering
    \includegraphics[width=.81\linewidth]{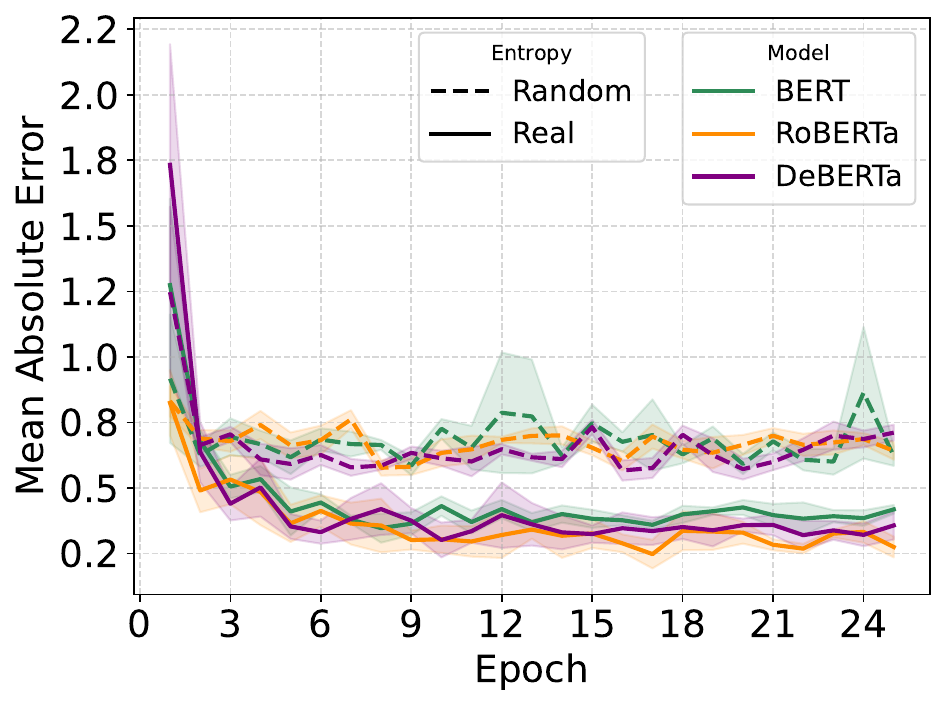}
    \caption{Mean Absolute Error values along training epochs on MPGT validation set across all models.}
    \label{fig:learning_curves_mae_singletask}

\end{figure}

Table~\ref{table:results_single_task} shows that models trained on real turn-transition entropy values consistently achieve lower MAE than those trained on randomly shuffled targets, across all datasets and backbone models. Figure~\ref{fig:learning_curves_mae_singletask} illustrates this behavior through the learning curves over 25 epochs on MPGT. Models trained on the true entropy signal converge after approximately six epochs and consistently outperform their shuffled-target counterparts. These results indicate that turn-transition entropy is a meaningful and learnable property of the dialogue context rather than an arbitrary continuous target. Therefore, we argue that its effectiveness in the multi-task setting is not solely attributed to a regularization effect, but also to the contextual information captured by the entropy signal itself.

\section{Multitask Learning on Random Entropy Targets}
\label{sec:ablation_effect}
We investigate the effect of using random targets as the auxiliary task in the multitask setting. Similar to our analysis in Section~\ref{sec:entropy_singletask}, we compare the proposed multitask setting against a control condition in which the entropy values are randomly shuffled across training instances. Table~\ref{table:ablation_random_effect} reports the difference in F1 score between the two settings ($\mathrm{F1}_{\text{random}} - \mathrm{F1}_{\text{real}}$). We observe that the differences are negative across all settings, showing that models trained with the true entropy values consistently outperform those trained with shuffled targets. In addition, we argue that incorporating the auxiliary task with random targets negatively impacts the performance of the main task. In particular, for both BERT and RoBERTa, the multitask models trained with shuffled targets achieve lower F1 scores than the baseline across all datasets. Overall, these findings demonstrate that the effectiveness of the proposed auxiliary task depends on the quality of the supervisory signal and provide further evidence that turn-transition entropy captures interaction patterns that can be leveraged for intent recognition.

\begin{table}[!t]
\small
\centering
\begin{tabular}{ l  c  c }

 \toprule

Model & MPGT & MIntRec2.0 \\ 

\midrule
BERT & -8.84 & -0.70 \\
RoBERTa & -5.53 & -0.29 \\
DeBERTa & -0.41 & -2.15 \\
\bottomrule

\end{tabular}
\caption{Absolute difference in macro-F1 score of the classification task when replacing the auxiliary regression targets with randomly shuffled values during multitask training ($\mathrm{F1}_{\text{random}} - \mathrm{F1}_{\text{real}}$). Negative values indicate a classification performance degradation when training with random regression targets.}

\label{table:ablation_random_effect}
\end{table}

\section{Target Variables Relationship Analysis}
\label{sec:emnlp_relationship_target_vars}
In this section, we assess the relationship between the auxiliary entropy target and the intent labels by conducting a Kruskal–Wallis test \cite{kruskal1952use}, which is appropriate for comparing a continuous non-Gaussian variable across multiple classes. The Kruskal–Wallis test is a non-parametric statistical test for comparing a continuous variable across $k$ independent groups, whose null hypothesis indicates no difference between groups. We provide the definition and mathematical background of the test in Appendix~\ref{app:app_targets_relationship_analysis}. Since the test provides only a $p$-value indicating statistical significance, without measuring an interpretable magnitude of differences, we also report the effect size $\epsilon^2$ (epsilon-squared), which ranges from 0 to 1. Values of $\epsilon^2$ closer to 1 indicate stronger associations between both variables. Further details on the effect size $\epsilon^2$ are provided in Appendix~\ref{app:app_targets_relationship_analysis}. This approach ensures that our conclusions are not only statistically significant but also practically meaningful. The results are summarized in Table~\ref{table:emnlp_relationship_target_vars}.

\begin{table}[ht]
\centering
\small
\begin{tabular}{lrrr}
\toprule
Dataset & $H$ & $p$-value & $\epsilon^2$ \\
\midrule
MPGT  & 51.57  & $7.11 \times 10^{-9}$  & 0.112 \\
MIntRec2.0 & 155.15 & $3.45 \times 10^{-19}$ & 0.021 \\
\bottomrule
\end{tabular}
\caption{Relationship analysis between the intent classes and the auxiliary variable. Kruskal-Wallis test results and effect sizes across datasets. $H$ corresponds to the Kruskal--Wallis statistic, and $\epsilon^2$ is the effect size}
\label{table:emnlp_relationship_target_vars}
\end{table}

We observe that the Kruskal–Wallis test rejects the null hypothesis that the entropy distributions are identical across intents for both datasets ($p<0.01$), indicating that the auxiliary variable is statistically associated with the intent labels. The corresponding effect sizes ($\epsilon^2$) suggest a stronger association for MPGT ($\epsilon^2 = 0.112$) than for MIntRec2.0 ($\epsilon^2 = 0.021$). Using the commonly adopted heuristic thresholds proposed by \citet{cohen1988statistical} and reviewed in \cite{tomczak2014need} for variance-explained effect sizes (small $\approx$ 0.01, medium $\approx$ 0.06, large $\approx$ 0.14), the association for MPGT can be regarded as moderate, whereas MIntRec2.0 exhibits a small association. We see that around 11\% and 2\% of the variability in the ranked entropy values are associated with the intent classes for MPGT and MIntRec2.0, respectively. Although the relationship is weaker in MIntRec2.0, both datasets show statistically significant dependence between the auxiliary target and the intent labels, supporting our intuition to conduct experiments on the use of the entropy regression task as an auxiliary objective. Finally, Figure~\ref{fig:emnlp_violin_entropy_by_class_mpgt} in Appendix~\ref{app:app_targets_relationship_analysis} shows the distributions of the turn-transition entropy values per class for MPGT, which illustrates the expected differences among intents.

\section{Conclusions}
\label{sec:conclusions}

We present a multitask learning method that captures turn taking dynamics as additional dialogue context for intent detection in multi-party conversations. Our auxiliary self-supervised task models the entropy on turn-taking patterns from speaker turn transitions to characterize dialogues. Our method shows improvement on intent detection systems over previous work approaches, using three model backbones over one private and two public multi-party dialogue corpora. We find that, such a turn-transition entropy can be modeled as a single task and thus does not only act as a regularizer in the multitask setting but is also a signal that provides information about the dialogue. We argue that our work, while performed for intent recognition, provides relevant insights for future work in other classification tasks involving multi-party conversations.


\section*{Limitations}
\label{sec:limitations}
While our approach does not impose constraints on the number of speakers, it assumes that speaker identities are available for each utterance, as is the case in the datasets used in our experiments. This requirement is naturally satisfied in scenarios where speaker diarization is controlled or where each participant owns a dedicated audio channel, e.g., video conferencing platforms. Furthermore, speaker identity is generally an easier label to obtain in real-world applications compared to the additional annotations required by alternative approaches, such as dialogue act labels or addressee identity, which typically demand costly expert annotation. Nevertheless, in scenarios where automatic diarization is applied, errors in speaker assignment may affect the quality of the turn-transition entropy signal. Studying the robustness of the turn-transition entropy to such noise is a promising direction for future work.



\section*{Ethical Considerations}
Our experiments use publicly available corpora, which have been curated prior to our work to prevent malicious actions. The internal corpus that we use in this work, has also been comprehensively curated to prevent malicious actions. Overall, the contributions presented in this paper are designed for constructive and ethical use, with no direct association with harmful social consequences. We do not find any potential risk on the release to the public of this work and the findings that we report.

The fine-tuning of all models developed in this work were executed on private infrastructure using a single NVIDIA A100 Tensor Core GPU of 80GB. The infrastructure has a carbon efficiency of 0.432 kgCO$_2$eq/kWh. The total time required for training across all experiments and analyses, including runs over 5 seeds per setting and hyperparameter tuning, was approximately 80 hours. Therefore, the total emissions are estimated at 13.82 kgCO$_2$eq. These estimations are
based on the Machine Learning Impact calculator\footnote{\url{https://mlco2.github.io/impact/}}
\cite{lacoste2019quantifying}.

\section*{Acknowledgments}
We warmly thank our anonymous reviewers for their time and valuable feedback. This work has been partially funded by the European Union's Horizon RIA research and innovation program under grant agreement No. 101189679 (ASTIR). This work also benefited from the FactoryIA supercomputer, financially supported by the Ile-de-France Regional Council.

\bibliography{acl_latex}

\appendix
\section{Utterance Properties Correlation Analysis}
\label{app:app_correlation_analysis}
We discussed in Section~\ref{sec:entropy} that there are utterance properties that might be correlated with the turn-transition entropy, which motivated the proposal of a normalized version of this auxiliary target. To investigate this relationship, we compute the Spearman correlation between the auxiliary variable (both normalized and unnormalized versions) and such utterance properties. The Spearman rank correlation coefficient measures the strength and direction of the monotonic association between two variables. For $n$ paired observations, it is defined as:
\begin{equation}
\rho_s = 1 - \frac{6\sum_{i=1}^{n} d_i^2}{n(n^2-1)},
\end{equation}
where $d_i$ is the difference between the ranks of the $i$-th pair of observations. The coefficient ranges from $-1$ to $1$, with values closer to $1$ ($-1$) indicating a stronger positive (negative) monotonic association. Table~\ref{table:emnlp_entropy_correlations}
reports the Spearman correlation coefficients among the variables, where an asterisk ($*$) indicates $p < 0.01$.

\begin{table}[ht!]
    \small
    \centering
    \begin{tabular}{llcc}
        \toprule
        Dataset & Property &
        Entropy & Norm. entropy \\
        \midrule
            & turn index      & 0.69\textsuperscript{*} & 0.46\textsuperscript{*} \\
            MPGT & context length  & 0.08   & 0.01   \\
            & speaker count   & 0.66\textsuperscript{*} & 0.20\textsuperscript{*} \\
        \midrule
            & turn index      & 0.77\textsuperscript{*} & 0.54\textsuperscript{*} \\
            MIntRec2.0 & context length  & 0.03   & 0.02   \\
            & speaker count   & 0.89\textsuperscript{*} & 0.24\textsuperscript{*} \\
        \bottomrule
    \end{tabular}
    \caption{Spearman correlations between entropy measures and utterance properties. An asterisk ($*$) indicates $p < 0.01$.}
    \label{table:emnlp_entropy_correlations}
\end{table}

We observe moderate to high correlations between the entropy values and the turn index and speaker count, on both corpora. However, we see that such relationship becomes weaker when the entropy values are normalized. Given the findings from Table~\ref{table:emnlp_entropy_correlations} and the results reported in Table~\ref{table:main_results}, we argue that normalizing the entropy variable prevents the model from capturing noise and colinear information from simple utterance properties such as turn index or speaker count. Hence, normalized entropy values better represent turn-taking dynamics without capturing such spurious information, as originally expected.

\section{Model Implementation Details}
\label{app:app_model_info}
In this appendix we provide model implementation details of our experiments.

\subsection{Language Models}
\label{app:language_models}
We use the large versions of uncased BERT\textsubscript, RoBERTa and DeBERTaV3 to build our baseline and multitask intent recognition models. All our fine-tuning runs were conducted on a single NVIDIA A100 Tensor Core GPU of 80GB. We used the \texttt{BertForSequenceClassification}, \texttt{RobertaForSequenceClassification}, \texttt{DebertaV2ForSequenceClassification} class from Hugging Face’s Transformers library \cite{wolf2020transformers} for sequence classification tasks. 

Work in \cite{qamar-etal-2023-speaking,he-etal-2021-speaker-turn} was implemented on the base version of RoBERTa. Thus, we reproduce their methods using RoBERTa\textsubscript{LARGE} for better comparability with our work. Additional experiments on prior research using RoBERTa\textsubscript{BASE} and their results are discussed in Appendix~\ref{app:comparison_roberta}.

\subsection{Large Language Models}
\label{app:large_language_models}
Figure~\ref{fig:emnlp_llm_baseline_prompt} shows the prompt template we use in our LLM baseline experiment on the intent recognition task.

\begin{figure}[htb!]
\centering
\begin{tcolorbox}[colback=gray!5!white, colframe=gray!40!black,
                  title=LLM Intent Classification Prompt,
                  fonttitle=\bfseries,
                  sharp corners=south, boxrule=0.4pt, width=0.95\linewidth]

**Task description**

Your task as an annotator is to assign a supported intent to the last utterance of a dialogue of multiple participants. Return the corresponding intent label. \\

**Authorized categories**

The supported intents and their definition are:

[\textcolor{blue}{INTENTS\_DEFINITIONS}] \\

**Previous utterances in the dialogue**

You have the following utterance history from multiple participants to understand the context of the dialogue:

[\textcolor{blue}{PREVIOUS\_UTTERANCES}] \\

**Expected output format**

Your response should contain the inferred intent between brackets as follows: [intent\_label] \\

Do not write anything else. \\

**Task**

The utterance to classify is shown below:

[\textcolor{blue}{UTTERANCE\_TO\_CLASSIFY}] \\

Predicted intent:

\end{tcolorbox}
\caption{Prompt template used on the LLM baseline for the intent classification task. Highlighted text in \textcolor{blue}{blue} varies among dataset examples.}
\label{fig:emnlp_llm_baseline_prompt}
\end{figure}

\subsection{Prior Work Implementation}
\label{app:prior_imp}

In our implementations of \cite{qamar-etal-2023-speaking,he-etal-2021-speaker-turn}, out-of-scope utterances are excluded from the training loss and evaluation metrics but retained as dialogue context. To preserve faithfulness to the baseline based on \citep{qamar-etal-2023-speaking}, hyperparameters are kept close to the original speaker-graph setup when specified. Unspecified hyperparameters are selected to follow relevant standards and support stability and memory usage. They are held identical across datasets, seeds, and encoder variants.


We use chunks of 128 utterances, dropout 0.5, weight decay \(5 \times 10^{-4}\),
learning rate \(10^{-4}\) for the encoder and \(10^{-3}\) for the graph head,
training for up to 100 epochs with early stopping after 10 non-improving
validation epochs. The learning rate is reduced by a factor of 0.1 after four
validation epochs without macro-F1 improvement. The batch size is fixed to 4 to
balance optimization stability and memory usage: smaller batches would yield  noisier gradient estimates, while larger batches increase the memory load of
128-utterance chunks processed through the RoBERTa, BiGRU, and RGAT modules
\cite{keskar2016large,mccandlish2018empirical}.

The maximum token length is set to 96 as a compromise between context preservation and transformer cost with shorter limits increasing the risk of truncating informative utterance content and longer limits the quadratic self-attention cost across all utterances in a chunk \cite{vaswani2017attention}. Gradient clipping is set to 1.0 as a conservative stabilization threshold for the recurrent component, following the standard use of norm clipping to mitigate exploding gradients in recurrent neural networks \cite{pascanu2013difficulty}.

\subsection{Our Baseline Implementation}
\label{app:baseline_imp}

Our baseline models implement an early stopping mechanism with a wait patience of 5. We use the macro-F1 score as the metric to monitor on the early stopping strategy. We use AdamW optimizer \cite{loshchilov2017decoupled} for optimizing the training objectives. Table~\ref{table:hyperparameters_baseline} shows the hyperparameter configuration we employ in our baseline experiments for all the models.


\begin{table}[!ht]
\centering
\small
\begin{tabular}{lrr} 
\toprule
hyperparameter & value MPGT & value MIntRec2.0\\
\midrule
eval\_monitor & macro F1  & macro F1 \\
train\_batch\_size & 16  & 16 \\
eval\_batch\_size & 16  & 16 \\
test\_batch\_size & 16  & 16 \\
wait\_patience & 5 & 5 \\
num\_train\_epochs & 60 & 60 \\
optimizer & AdamW & AdamW \\
warmup\_proportion & 0.1 & 0.1 \\
lr & 1e-5 & 1e-5 \\
seeds & 0-4 & 0-4 \\

\bottomrule
\end{tabular}
\caption{Set of hyperparameters used on BERT, RoBERTa, and DeBERTa fine-tuning baseline experiments.}
\label{table:hyperparameters_baseline}
\end{table}

\subsection{Multi-task Method Implementation}
\label{app:multitask_imp}

In addition to the early stopping mechanism applied on the main task (i.e., intent classification), we also set a less tolerant early stopping strategy on the auxiliary task (i.e., entropy regression). We argue that the auxiliary task acts in part as a regularizer during the model fine-tuning, which negatively impacts the main task performance after some training iterations. Hence, we define a patience of 2 epochs for the auxiliary task, which is disconnected from the total loss computation when the early stopping is activated. To do so, we monitor the auxiliary task on the MAE. For the main task, we keep a wait patience of 5. By doing so, we force the model to solely optimize the intent classification loss in the last epochs of the model fine-tuning.

Our proposed multi-task methods includes two additional hyperparameters to the set of hyperparameters used in our baseline experiments (see Section~\ref{sec:classification} and Appendix~\ref{app:baseline_imp}). First, we add the hyperparameter $\lambda$, which controls the weight of the auxiliary loss in the average loss computation on the multi-task setting. Moreover, we include the layer number in our architecture where we connect the regression head. Both hyperparameters are optimized using Optuna. Table~\ref{table:optuna_search_space} shows the search spaces we define for both hyperparameters. Table~\ref{table:hyperparameters_multitask} shows the final hyperparameter sets that we use in the multi-task setting experiments, including the optimized hyperparameters $\lambda$ and \textit{l}.

\begin{table}[t]
\centering
\begin{tabular}{lcc}
\toprule
hyperparameter & search range & step \\
\midrule
weight $\lambda$ & $[0.01,\,0.20]$ & $0.01$ \\
layer $l$ & $[3,\,21]$ & $1$ \\
\bottomrule
\end{tabular}
\caption{Optuna hyperparameter search space for the average weight $\lambda$ and layer number \textit{l} on the multi-task architecture.}
\label{table:optuna_search_space}
\end{table}

\begin{table}[!ht]
\centering
\small
\begin{tabular}{lrr} 
\toprule
hyperparameter & value MPGT & value MIntRec2.0\\
\midrule
BERT &  & \\
eval\_monitor\_main & macro F1  & macro F1 \\
eval\_monitor\_aux & MAE  & MAE \\
train\_batch\_size & 16  & 16 \\
eval\_batch\_size & 16  & 16 \\
test\_batch\_size & 16  & 16 \\
wait\_patience\_main & 5 & 5 \\
wait\_patience\_aux & 2 & 2 \\
num\_train\_epochs & 60 & 60 \\
optimizer & AdamW & AdamW \\
warmup\_proportion & 0.1 & 0.1 \\
lr & 5e-6 & 1e-5 \\
seeds & 0-4 & 0-4 \\
weight $\lambda$ & 0.01 & 0.13 \\
layer \textit{l} & 18 & 9 \\

\midrule

RoBERTa &  & \\
eval\_monitor & macro F1  & macro F1 \\
eval\_monitor\_aux & MAE  & MAE \\
train\_batch\_size & 16  & 16 \\
eval\_batch\_size & 16  & 16 \\
test\_batch\_size & 16  & 16 \\
wait\_patience\_main & 5 & 5 \\
wait\_patience\_aux & 2 & 2 \\
num\_train\_epochs & 60 & 60 \\
optimizer & AdamW & AdamW \\
warmup\_proportion & 0.1 & 0.1 \\
lr & 5e-6 & 1e-5 \\
seeds & 0-4 & 0-4 \\
weight $\lambda$ & 0.07 & 0.08 \\
layer \textit{l} & 18 & 16 \\

\midrule

DeBERTa &  & \\
eval\_monitor & macro F1  & macro F1 \\
eval\_monitor\_aux & MAE  & MAE \\
train\_batch\_size & 16  & 16 \\
eval\_batch\_size & 16  & 16 \\
test\_batch\_size & 16  & 16 \\
wait\_patience\_main & 5 & 5 \\
wait\_patience\_aux & 2 & 2 \\
num\_train\_epochs & 60 & 60 \\
optimizer & AdamW & AdamW \\
warmup\_proportion & 0.1 & 0.1 \\
lr & 5e-6 & 1e-5 \\
seeds & 0-4 & 0-4 \\
weight $\lambda$ & 0.01 & 0.07 \\
layer \textit{l} & 6 & 15  \\

\bottomrule
\end{tabular}
\caption{Set of hyperparameters used on BERT, RoBERTa, and DeBERTa fine-tuning multi-task method experiments after Bayesian optimization.}
\label{table:hyperparameters_multitask}
\end{table}


\section{Dataset Statistics}
\label{app:dataset_info}
Table \ref{table:dataset_stats} shows the number of examples per dataset split. All corpora contain both in-scope and out-of-scope (OOS) intents. That means that there are many utterances that are not associated to any specific intent. Our classification task is defined over in-scope examples. MPGT, MIntRec2.0 and MEETInt contain 22\%, 38\% and 0.5\% of in-scope examples to classify, respectively. Despite we do not use OOS examples as target utterances to classify, we retain them when they are part of the previous utterances as they provide contextual information.

\begin{table}[htb]
\footnotesize
\centering
\renewcommand{\arraystretch}{1.5}  
\begin{tabular}{lrrrr} 
\toprule

 & train & dev & test & labels \\ 
\midrule
MPGT & .5K & .1K & .2K & 8 \\
MIntRec2.0 & 9.9K & 1.8K & 3.2K & 30 \\
MEETInt & 81.9K & 21.0K & 31.8K & 12  \\

\bottomrule
\end{tabular}
\caption{Number of examples per dataset split and number of labels on each corpus.}
\label{table:dataset_stats}
\end{table}


\section{Speaker Embedding Experiments}
\label{app:speaker_emb_experiments}
We perform additional experiments in order to determine a strong baseline method for our work. In particular, we study different input representations with a fixed architecture over the same backbone models that we describe in Section~\ref{sec:experiments}. The input representations are described as follows:
\begin{enumerate}
    \item \textbf{w/o context:} Each example corresponds to the target utterance to classify without prepending previous utterances as context.
    \item \textbf{only target speaker:} Following \cite{zhang2024mintrec2}, each example corresponds to the target utterance to classify and the previous utterances produced by the speaker of the target utterance to classify (i.e. targer speaker) as context.
    \item \textbf{all speakers w/o speaker emb.:} Each example is represented by the target utterance to classify and the utterances previously spoken by any speaker.
    \item \textbf{all speakers w. speaker emb.:} Each example is represented by the target utterance to classify and the utterances previously spoken by any speaker. In addition, we add a speaker embedding representation that indicates what utterances have been spoken by the target speaker.
\end{enumerate}

\begin{table*}[!t]
\small
\centering
\begin{tabular}{c l c c | c c }
 \toprule

 & & \multicolumn{2}{c|}{MPGT} & \multicolumn{2}{c}{MIntRec2.0}\\

Model & Input Representation & F1 & Acc. & F1 & Acc. \\ 

\midrule

\multirow{4}{*}{BERT} & w/o context & 86.84 & 95.04 & 53.08 & 58.67  \\

 & only target speaker & 87.92 & \textbf{95.79} & 55.18 & 60.45  \\

 & all speakers w/o speaker emb. & 87.69 & 94.69 & 54.74 & 59.94  \\

 & all speakers w. speaker emb. & \textbf{90.65} & 95.49 & \textbf{55.36} & \textbf{60.57}  \\

\midrule
\midrule

\multirow{4}{*}{RoBERTa} & w/o context & 81.85 & 93.83 & 55.44 & 59.85  \\

 & only target speaker & 83.49 & 94.59 & 55.55 & 60.82  \\

 & all speakers w/o speaker emb. & 82.62 & \textbf{95.04} & 55.32 & 60.52  \\

 & all speakers w. speaker emb. & \textbf{83.68} & 94.89 & \textbf{56.87} & \textbf{61.24}  \\

\midrule
\midrule

\multirow{4}{*}{DeBERTa} & w/o context & 81.79 & 91.74 & 53.43 & 58.67  \\

 & only target speaker & 81.80 & 91.58 & 54.79 & 59.73.  \\

 & all speakers w/o speaker emb. & \textbf{82.30} & \textbf{95.34} & 55.41 & \textbf{60.41}  \\
 
 & all speakers w. speaker emb. & 77.95  & 92.93 & \textbf{55.63} & 60.27 \\

\bottomrule

\end{tabular}
\caption{Performance evaluation on the intent classification task using different input representation strategies. Scores correspond to the average macro F1 and accuracy over 5 runs. Scores in \textbf{bold} highlight \textbf{the best} performance on the same backbone model.}

\label{table:speaker_embeddings_results}
\end{table*}

For these experiments, we use the same hyperparameter values reported in Table~\ref{table:hyperparameters_baseline}, except that the models are fine-tuned over 40 epochs and we use early stopping with a patience of 3 epochs. We observe in Table~\ref{table:speaker_embeddings_results} that the approach proposed by \citet{zhang2024mintrec2} outperforms adding all speakers previous utterances without speaker embeddings on BERT and RoBERTa. Nevertheless, our adding our proposed speaker embeddings shows performance gains. Overall, we see that including utterances produced by all speakers with speaker embeddings provides better performance than any other strategy on both models. On the other hand, we see that the best strategy on DeBERTa is the representation that includes all speaker context utterances without speaker embeddings. Similar to the results on BERT and RoBERTa, solely providing target speaker context utterances is not the best strategy on DeBERTa. For consistency and better comparability across models, we select the  \textbf{all speakers w. speaker emb.} strategy as the input representation for both our baseline and our multi-task learning method.

\section{Comparison of RoBERTa\textsubscript{BASE} and RoBERTa\textsubscript{LARGE} on Prior Works}
\label{app:comparison_roberta}
The experiments on prior works \cite{he-etal-2021-speaker-turn,qamar-etal-2023-speaking} that we reproduce and report in Section~\ref{sec:results} are conducted on RoBERTa\textsubscript{LARGE}, since our methods and other baseline approaches are implemented on the large versions of pre-trained models. Our aim is to keep fair comparisons among methods. Nevertheless, results reported in \cite{he-etal-2021-speaker-turn,qamar-etal-2023-speaking} are obtained from experiments on the base versions of RoBERTa. Therefore, we make additional experiments on prior work implemented with RoBERTa\textsubscript{BASE}. Table \ref{table:comparison_roberta} shows the performance comparison between the two versions of RoBERTa. Note that, while the results on the base version of RoBERTa are higher, such scores are still lower than our proposed method evaluation presented in Table~\ref{table:main_results} from Section~\ref{sec:results}.

\begin{table*}[!t]
\small
\centering
\begin{tabular}{c l c c | c c }
 \toprule

 & & \multicolumn{2}{c|}{MPGT} & \multicolumn{2}{c}{MIntRec2.0}\\

Model & Method & F1 & Acc. & F1 & Acc. \\ 

\midrule

\multirow{2}{*}{RoBERTa\textsubscript{BASE}} & Turn Modeling \cite{he-etal-2021-speaker-turn} & 82.00  & 94.26 & 49.76 & 57.68 \\

 & Turn Aware Speaker Graph \cite{qamar-etal-2023-speaking} & 83.30  & 95.19 & 52.31 & 57.70 \\

\midrule

\multirow{2}{*}{RoBERTa\textsubscript{LARGE}} & Turn Modeling \cite{he-etal-2021-speaker-turn} & 73.64  & 92.92 & 51.40 & 56.40 \\

 & Turn Aware Speaker Graph \cite{qamar-etal-2023-speaking} & 81.28  & 94.29 & 52.82 & 57.92 \\

\bottomrule

\end{tabular}
\caption{Performance evaluation on the intent classification task. Comparison between RoBERTa\textsubscript{BASE} and RoBERTa\textsubscript{LARGE} on prior work baselines. Scores correspond to the average macro F1 and accuracy over 5 runs.}

\label{table:comparison_roberta}
\end{table*}

\section{Target Variables Relationship Analysis Details}
\label{app:app_targets_relationship_analysis}
\begin{figure*}[t]
    \centering
    \includegraphics[width=1.0\linewidth]{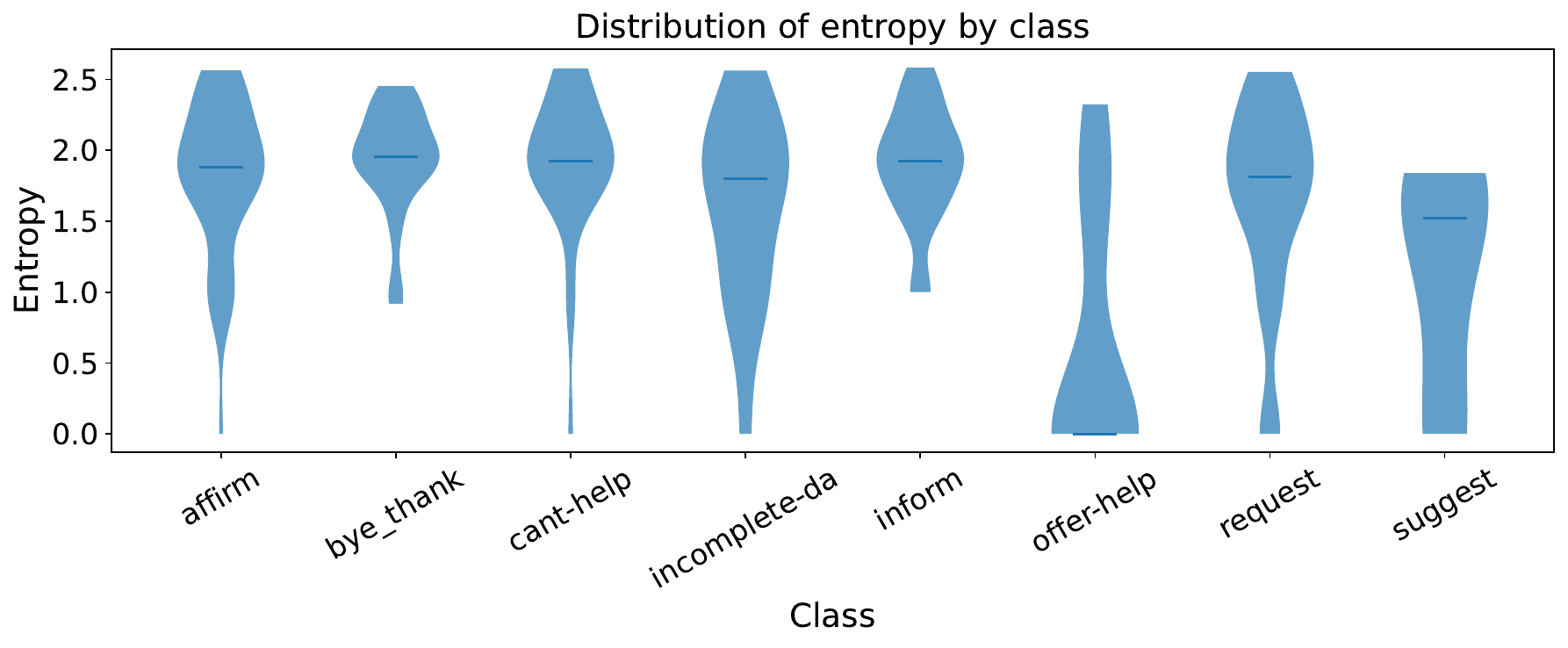}
    \caption{Distribution of the turn-transition entropy values across classes for MPGT.}
    \label{fig:emnlp_violin_entropy_by_class_mpgt}

\end{figure*}

In this appendix, we provide further details of the statistical relationship analysis described in Section~\ref{sec:emnlp_relationship_target_vars}. The Kruskal--Wallis test is a non-parametric statistical test for comparing a continuous variable across $k$ independent groups, whose null hypothesis indicates no difference between groups. Rather than operating on raw values, the test converts observations to their ranks.  Let $N$ denote the total number of observations, $n_i$ the sample size for group $i$, and $R_i$ the sum of ranks assigned to observations in group $i$. The Kruskal--Wallis test statistic $H$ is computed as:
\begin{equation}
H = \frac{12}{N(N+1)} \sum_{i=1}^{k} \frac{R_i^2}{n_i} - 3(N+1),
\end{equation}
where $H$ approximately follows a chi-squared distribution with $k-1$ degrees of freedom under the null hypothesis of no difference between groups. Unlike parametric alternatives such as ANOVA \cite{fisher1930statistical}, this rank-based approach does not assume normality of the data. Thus, given the peaks around 0 and 1 as well as the tails observed in Figure~\ref{fig:entropy_distributions}, the lack of normality assumption is essential for our analysis. However, the Kruskal--Wallis test provides only a $p$-value indicating statistical significance, without directly measuring the practical magnitude of differences. To address this limitation, we also report the effect size $\epsilon^2$ (epsilon-squared), computed as:
\begin{equation}
\epsilon^2 = \frac{H - k + 1}{N - k},
\end{equation}
where $H$ is the test statistic, $k$ is the number of groups being compared, and $N$ is the total sample size. The effect size $\epsilon^2$ ranges from 0 to 1, with values closer to 1 indicating stronger associations between the grouping variable and the continuous outcome. Finally, Figure~\ref{fig:emnlp_violin_entropy_by_class_mpgt} shows the distributions of the turn-transition entropy values per class for MPGT, which illustrates the expected differences among intents.

\end{document}